\documentclass{article}

\usepackage[nonatbib, final]{nips_2016}

\usepackage[utf8]{inputenc} 
\usepackage[T1]{fontenc}    
\usepackage{hyperref}       
\usepackage{url}            
\usepackage{booktabs}       
\usepackage{amsfonts}       
\usepackage{nicefrac}       
\usepackage{microtype}      
\usepackage{graphicx}
\usepackage{caption}
\usepackage{cite}

\title{Block Neural Network Avoids Catastrophic Forgetting When Learning Multiple Task}

\author{
  Guglielmo Montone\\
  Laboratoire Psychologie de la Perception\\
  Universit\'e Paris Descartes\\
  75006 Paris, France\\
  \texttt{montone.guglielmo@gmail.com} \\
  \And
  J.Kevin O'Regan\\
  Laboratoire Psychologie de la Perception\\
  Universit\'e Paris Descartes\\
  75006 Paris, France\\
  \texttt{jkevin.oregan@gmail.com} \\
  \AND
  Alexander V. Terekhov\\
  Laboratoire Psychologie de la Perception\\
  Universit\'e Paris Descartes\\
  75006 Paris, France\\
  \texttt{avterekhov@gmail.com} \\
}

\begin{document}

\maketitle

\begin{abstract}
In the present work we propose a Deep Feed Forward network architecture which can be trained according to a sequential learning paradigm, where tasks of increasing difficulty are learned sequentially, yet avoiding \textit{catastrophic forgetting}. The proposed architecture can re-use the features learned on previous tasks in a new task when the old tasks and the new one are related. The architecture needs fewer computational resources (neurons and connections) and less data for learning the new task than a network trained from scratch
\end{abstract}

\section{Introduction}
Two recently suggested architectures, the block neural network \cite{terekhov2015knowledge, montone2015usefulness} and the progressive neural network \cite{rusu2016progressive}, tested respectively in a supervised learning paradigm and a reinforcement learning paradigm have shown impressive results in \textit{multi-task learning}. The block neural network is created by training several Deep Feed Forward networks (DNNs) on different tasks. The networks are then connected using new neurons and connections, forming a bigger network that is trained on a new task by allowing just the new added connections to be updated. Block neural networks and progressive neural networks have both been shown to benefit from the advantages of transfer learning. Whereas in the past different forms of \textit{pre-training} \cite{erhan2010does, mesnil2012unsupervised} and \textit{multi-task learning} \cite{caruana1998multitask} have also achieved this, block neural networks and progressive networks do so without suffering from the disadvantage of \textit{catastrophic forgetting} of old tasks in the case of pre-training and the necessity of a persistent reservoir of data for the multi-task learning. In this paper, after quickly revisiting the block network architecture, we propose a set of binary classification tasks and show that the block architecture learns more simply (the network needs less computational resources: neurons and connections) and more quickly (the train set can be much smaller) than a network trained from scratch.
\section{Merging DNNs}
\begin{figure}[h]
  \centering
  \includegraphics[width=\textwidth, height=0.13\textheight]{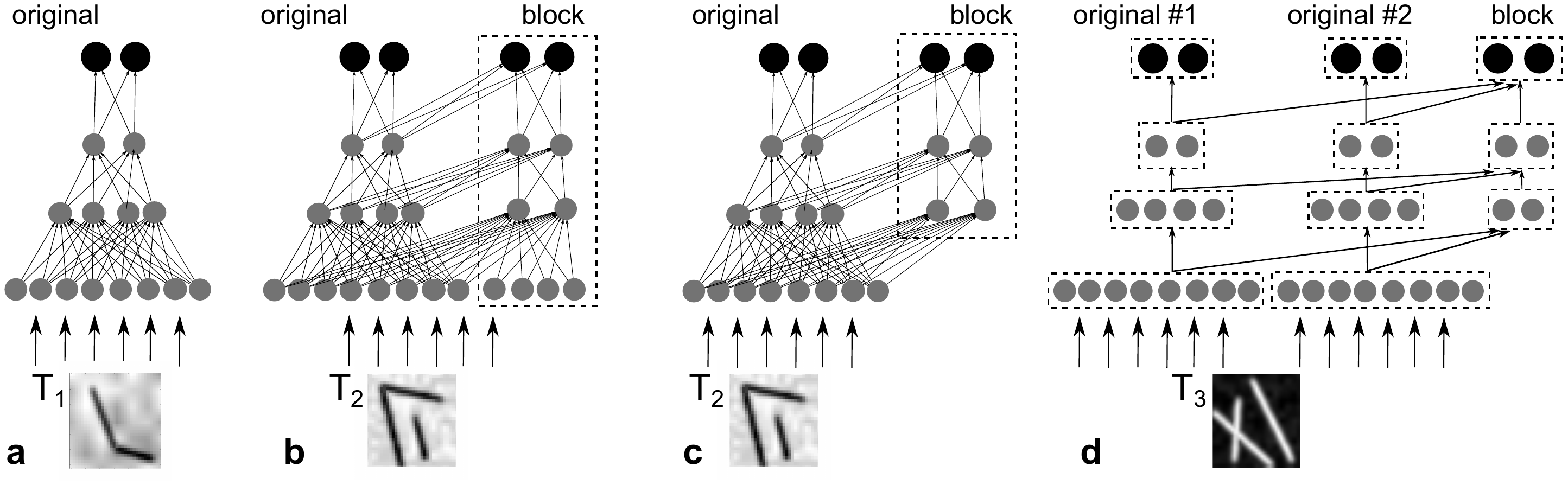}
  \caption{(a) The architecture is built by adding a block of neurons with three hidden layers to one base model. (b) Adding a block of neurons with two hidden layers to one base model. (c) Adding a block of neurons with one hidden layer to one base model. (d) Adding a block of neurons to two base models. The dashed boxes indicate the layers of the two base models and the block of neurons added. An arrow connecting two boxes indicates that all the neurons in the first box are connected to all the neurons in the second box.}
  \label{figure:blockArchitecture}
\end{figure}
We defined a set of tasks $T_1, \dots, T_M$ and trained a DNN $N_1, \dots, N_M$ (\textit{base models}) on each task. After the first training phase, we used some of the trained networks, say $N_1, \dots, N_m$, to build a \textit{block architecture} that was then trained on one of the remaining tasks, say $T_{m+}$. The block architecture was formed by adding a set of new neurons (\textit{block neurons}) to the previously trained networks $N_1 ,\dots,N_m$. The block neurons were connected to the base models as follows: the first hidden layer of the block neurons received the input for the task $T_{m+1}$ . The same input was sent to all networks $N_1, \dots, N_m$ . The second hidden layer was fully connected to both the first hidden layer of the block neurons and the first hidden layer of each network $N_1 ,\dots, N_m$ . This pattern was repeated for all the layers. This architecture was tested with two variations. In the two variations respectively the first and the second layer of the block neurons were removed. When training on the task $T_{m+1}$ \textit{none of the parameters in the base model networks was allowed to change}. Figure \ref{figure:blockArchitecture} provides a representation of the block neural network.

\section{The tasks}
We used six binary classification tasks, which the networks were trained on. The tasks all
involved the concepts of line and angle. We wished to show that the networks $N_1, \dots, N_m$, when
trained on such tasks, would develop features that could be reused by the block architecture to solve
another task involving the same concepts.
\begin{figure}[h!]
  \centering
  \includegraphics[width=\textwidth, height=0.07\textheight]{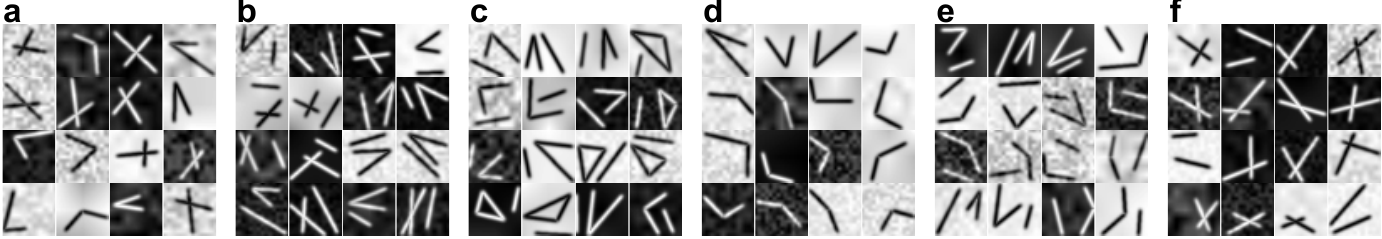}
  \caption{Examples of stimuli: (a) \textit{ang\_crs}; (b) \textit{ang\_crs\_ln}; (c) \textit{ang\_ tri ln}; (d) \textit{blt\_srp}; (e) \textit{blt\_srp\_ln}; (f) \textit{crs ncrs}}
\label{figure:stimuli}
\end{figure}
In each task the stimuli were gray scale images, $32 \times 32$ pixels in size. Each image contained two to four line segments, each at least 13 pixels long (30\% of the image diagonal). The segments were white on a dark random background or black on a light random background.  The 6 tasks were (see examples in figure \ref{figure:stimuli}):

\vspace{0.3pt}
\textit{ang\_crs}: requires classifying the images into those containing an angle (between $20^{\circ}$ and $160^{\circ}$) and a pair of crossing line segments (the crossing point must lie between 20\% and 80\% along each segment’s length).

\vspace{0.3pt}
\textit{ang\_crs\_ln}: the same as \textit{ang\_crs}, but has an additional line segment crossing neither of the other line segments.

\vspace{0.3pt}
\textit{ang\_tri\_ln}: distinguishes between images containing an angle (between $20^{\circ}$ and $160^{\circ}$) and a triangle (with each angle between $20^{\circ}$ and $160^{\circ}$ ); each image also contains a line segment crossing neither angle nor triangle.

\vspace{0.3pt}
\textit{blt\_srp}: requires classifying the images into those having blunt (between $100^{\circ}$ and $160^{\circ}$ ) and those having sharp (between $20^{\circ}$ and $80^{\circ}$ ) angles in them.

\vspace{0.3pt}
\textit{blt\_srp\_ln}: the same as \textit{blt\_srp}, but has an additional line segment, crossing neither of the line segments forming the angle.

\vspace{0.3pt}
\textit{crs\_ncrs}: distinguishes between a pair of crossing and a pair of non-crossing lines (the crossing
point must lay between 20\% and 80\% of each segment length).

\section{Results}
\begin{table}[t]
  \caption{Original network results}
  \label{table: Original network}
  \centering
  \begin{tabular}{lcc}
    Condition  & 200-100-50 (300K params) & 60-40-20 (65K params)\\
    \midrule
    \textit{ang\_crs}     &   5.5(5.4-5.9)    & 9.4(8.9-9.8)\\
    \textit{ang\_crs\_ln} &   13.6(12.5-15.2) & 18.3(16.7-18.8)\\
    \textit{ang\_tri\_ln} &   6.1(5.5-6.8)    & 11.4(10.6-14.0)\\
    \textit{blt\_srp}     &   2.0(1.8-2.3)    & 3.7(3.4-4.2)\\
    \textit{blt\_srp\_ln} &   6.5(6.4-6.9)    & 12.5(11.6-14.1)\\
    \textit{crs\_ncrs}    &   2.8(2.3-2.9)    & 4.5(4.1-5.2)\\
    \bottomrule
  \end{tabular}
\end{table}
In this section, we first report the results obtained by training a DNN on each of the previously described tasks. Then we report the results of training different block neural networks on the same tasks. The number of possible architectures that can be built by changing the base models, the number of block neurons and the task on which the block network is trained, is very large, and exploring all possibilities was not feasible. A more detailed analysis of the configurations tried can be found in our previous studies\cite{terekhov2015knowledge, montone2015usefulness}. Here we summarize the results obtained with two kinds of block network architectures that are particularly interesting because they are obtained by adding a very small number of block neurons. Moreover in this paper we focus on the ability of such architectures to learn using a much smaller dataset. We will in fact present the performance obtained by several block architectures when such architectures are trained on a dataset of almost half the size of the dataset used for training a network from scratch.\\
The performance of the networks was evaluated by computing the percentage of misclassified samples on the test dataset. Each architecture was trained five times, randomly initializing its weights. The mean performance over the five repetitions and the best and worst performance are reported in the tables.
\begin{figure}[h]
  \centering
  \includegraphics[scale=0.2]{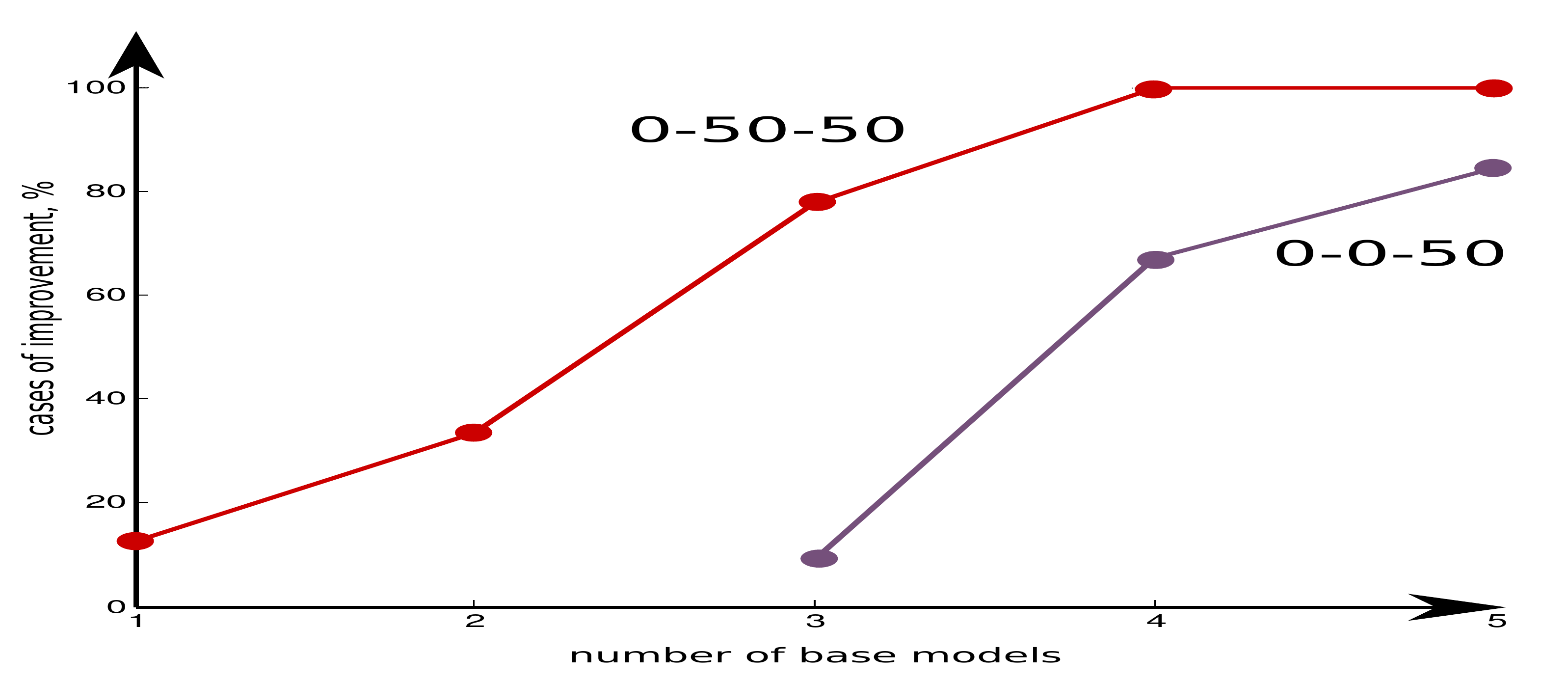}
  \caption{Percentage of block architectures outperforming a network trained from scratch as a function of the number of base models present in the block architecture}
  \label{figure: cumulative results}
\end{figure}

\subsection*{Original Network}
Prior to building block architectures, we trained a DNN on each task. The networks used were of type NN-200-100-50, with 200, 100, and 50 nodes in the first, second, and third layers, respectively. Networks of this type were used as base models for all of the block networks. The percentages of misclassified test examples for these networks are shown in table \ref{table: Original network} together with the results for another architecture, namely NN-60-40-30. Such networks had approximately the same number of parameters (weight of the networks) as some of the block networks, making interesting performance comparisons possible. The networks were trained on datasets with $350,000$ examples.

\subsection*{Block Architecture}
In figure \ref{figure: cumulative results} we present the percentage of block networks outperforming a network trained from scratch as a function of the number of base models present in the block network. Here we focus on two kinds of block architecture, namely BA-0-50-50 and BA-0-0-50. The architecture BA-0-50-50 (BA-0-0-50) is obtained by connecting the base models to a DNN  with 0(0) units in the first hidden layer, 50(0) units in the second hidden layer, and 50 units in the third hidden layer. The plots in figure \ref{figure: cumulative results} were obtained as follow. We built several instantiations of the two kinds of block architectures by using randomly selected base models. Each architecture was then trained on each of the tasks if the task was not used to train any of its base models. For example a block network built using the base model trained on \textit{blt\_srp} and \textit{ang\_crs} was trained on all the other tasks excepts those two. The performance obtained by each block network was compared with the performance obtained with a network (NN-200-100-50) trained from scratch on the same task. 
\begin{table}[t]
  \caption{Block architecture with four base models. Dataset of $200.000$ stimuli}
  \label{table: Adding blocks to four base models}
  \centering
  \begin{tabular}{lcc}
    Condition  & BA-0-50-50 (60K params) & BA-0-0-50 (5K params)\\
    \midrule
    \textit{ang\_crs (ang\_tri\_ln+crs\_ncrs+blt\_srp+blt\_srp\_ln)}       &   \textbf{5.0(4.8-5.2)}    & 5.8(5.4-6.3)\\
    \textit{ang\_crs (ang\_tri\_ln+ang\_crs\_ln+crs\_ncrs+blt\_srp\_ln)}   &   \textbf{4.3(4.0-4.5)}    & \textbf{4.6(4.0-5.0)}\\
    \textit{ang\_crs (ang\_tri\_ln+crs\_ncrs+blt\_srp\_ln+ang\_crs\_ln)}   &   \textbf{4.3(4.1-4.8)}    & \textbf{4.7(4.3-5.5)}\\
    \textit{ang\_crs\_ln (ang\_tri\_ln+crs\_ncrs+blt\_srp\_ln+ang\_crs)}   &   \textbf{10.7(10.4-11.3)}   & \textbf{12.0(11.5-12.4)}\\
    \textit{ang\_crs\_ln (ang\_tri\_ln+crs\_ncrs+blt\_srp+blt\_srp\_ln)}   &   \textbf{12.4(12.0-12.6)} & 15.1(14.6-15.5)\\
    \textit{blt\_srp (ang\_crs+ang\_tri\_ln+crs\_ncrs+blt\_srp\_ln)}       &   \textbf{1.2(1.1-1.4)}    & \textbf{1.4(1.3-1.5)}\\
    \textit{blt\_srp (ang\_crs\_ln+ang\_tri\_ln+crs\_ncrs+ang\_crs)}       &   \textbf{1.8(1.7-2.0)}    & 2.1(1.7-2.4)\\
    \textit{blt\_srp\_ln (ang\_crs\_ln+ang\_tri\_ln+crs\_ncrs+ang\_crs)}   &   \textbf{6.4(6.3-6.6)}    & 9.7(9.2-10.6)\\
    \textit{blt\_srp\_ln (ang\_crs\_+ang\_tri\_ln+crs\_ncrs+ang\_crs\_ln)} &   \textbf{6.5(6.3-6.8)}    & 9.8(9.4-10.3)\\   
    \bottomrule
  \end{tabular}
\end{table}
The percentage of times the block network obtained a better score was evaluated. The plot in the figure clearly shows an increase in the performance of the block network as the number of base models grows. On the one hand this result was expected simply because the bigger the number of base models, the more parameters are trained; on the other hand it is important to stress that the number of parameters trained on the block network is in any case much smaller than that of a network trained from scratch. The architecture BA-0-50-50 with five base models, for example, has about 60K parameters compared to the 300K of the network 200-100-50.
\begin{table}[h]
  \caption{Block network with five base models. Dataset of $200.000$ examples}
  \label{table: Adding blocks to five base models}
  \centering
  \begin{tabular}{lcc}
    Condition  & BA-0-50-50 (75K params) & BA-0-0-50 (25K params)\\
    \midrule
    \textit{ang\_crs} (all model used except \textit{ang\_crs})         &   \textbf{4.3(4.0-4.7)}    & \textbf{4.4(3.9-4.7)}\\
    \textit{ang\_crs\_ln} (all model used except \textit{ang\_crs\_ln}) &   \textbf{10.6(10.4-10.8)}    & \textbf{11.7(11.2-12.1)}\\
    \textit{blt\_srp} (all model used except \textit{blt\_srp})         &   \textbf{1.2(0.9-1.9)}    & \textbf{1.4(1.1-1.8)}\\
    \textit{blt\_srp\_ln} (all model used except \textit{blt\_srp\_ln}) &   \textbf{5.6(5.2-5.9)}    & 7.2(6.8-8.0)\\
    \textit{crs\_ncrs} (all model used except \textit{crs\_ncrs})       &   \textbf{1.2(1.0-1.3)}    & \textbf{1.2(1.0-1.3)}\\
    \textit{ang\_tri\_ln} (all model used except \textit{ang\_tri\_ln}) &   \textbf{5.8(5.7-6.0)}    & 8.6(8.3-9.0)\\   
    \bottomrule
  \end{tabular}
\end{table}
In table \ref{table: Adding blocks to four base models} and table \ref{table: Adding blocks to five base models} we show the performance of the block network when the network is trained on a dataset of $200,000$ examples, almost half of the size of the dataset used to train the network NN-200-100-50. The percentages of misclassified test examples for block architectures with four and five base models are presented in the tables. The architectures that performed better than (or equal to) the network NN-200-100-50, which was trained from scratch, are shown in bold. In these tables, the tasks on which the block architectures were trained are listed together with the tasks on which the base models were trained (in parentheses).

\section{Conclusions}
The block architecture proves to be a very effective solution for approaching the problem of multi-task learning in DNN. The architecture can be a first step toward the construction of DNN architectures which, in an unsupervised fashion, are able to profit from training on prior tasks when learning a new task. 

\subsubsection*{Acknowledgments}
This work was funded by the ERC proof of concept grant number 692765 "FeelSpeech"

\newpage
\bibliography{biblio}{}
\bibliographystyle{plain}

\end{document}